# The Deontic Gap: Large Language Models and the Modal Language of Obligation

**Daniel Hart[1], Sarah Allred[2], Joseph Abbas[3], and Morenike Alugo[3]**
[1] Graduate School of Applied and Professional Psychology, Rutgers University, New Brunswick, NJ, USA [2] Department of Psychology, Rutgers University–Camden, Camden, NJ, USA [3] Prevention Science Program, Rutgers University–Camden, Camden, NJ, USA

**Corresponding author:** Daniel Hart, Graduate School of Applied and Professional Psychology, Rutgers University, New Brunswick, NJ, USA. Email: daniel.hart@rutgers.edu
**Preprint.** This manuscript is under review at *Journal of Pragmatics*. This version has not been peer reviewed. Analysis code and derived data are openly available at https://doi.org/10.17605/OSF.IO/YCJWH.

## Abstract
Modal auxiliaries such as *must*, *should*, and *have to* mark necessity and obligation within the contexts of speaker authority and interpersonal stance. We examine whether large language models (LLMs) reproduce contemporary human patterns of deontic modal usage. Across three primary corpora, an external benchmark, two controlled replications, and a naturalistic eleven-model replication, AI-generated text consistently underuses positive deontic modals (*must*, *should*, *have to*, *had to*) relative to contemporary humans. Historical comparison with the Google Books Ngram corpus (1920-2022), used as a heuristic calibration against the published-prose record, shows that AI modal frequencies fall within the range of formal published English, whereas contemporary human modal rates in informal digital contexts often exceed twentieth-century book baselines. Phrase-level decomposition shows that the AI–human modal gap is concentrated in constructions central to interpersonal stance (*should*, *have to*, *had to*), while AI matches or exceeds humans on *need to* in instructional and question-answering contexts but not in persuasive student writing, indicating that the modal profile is genre-conditional. The findings suggest that LLM modal usage reflects the formal written resources on which these models were trained, while underusing the modal constructions through which contemporary human writers mark immediate, interpersonal obligation.

## 1. Introduction
"You *should* brush your teeth." "You *must* return library books." "You *have to* behave in class." Each sentence issues an obligation, yet the three obligations differ in force, in the authority they invoke, and in the relationship they presuppose between speaker and listener. Modal verbs of this kind, which linguists call deontic modals because they mark duty and requirement rather than mere possibility, allow speakers and writers to calibrate what is required, recommended, expected, or simply permitted (Knobe & Szabo, 2013;

Phillips & Kratzer, 2024). In doing so, they also position people in social space. A parent who says “you have to” claims a different footing than an official notice announcing that “patrons must.” The choice among deontic modals is therefore never only a grammatical choice. It is a choice about how obligation itself is to be represented.

These choices have consequences for judgment and behavior. Kuang and Bicchieri (2024) showed experimentally that injunctive formulations of a norm, telling people what they should do, produce stronger shifts in norm perception and compliance than descriptive formulations of the same content. A broader literature on moralized language finds that small lexical differences alter how strongly a message registers as obligatory and how readily it spreads through social networks (Brady et al., 2017; Capraro & Vanzo, 2019). Moreover, work in cognitive science indicates that deontic judgment and modal reasoning draw on shared and fundamental machinery of human thought (Phillips & Cushman, 2017; Shtulman & Tong, 2013). Modal language, in short, is a central vehicle through which people encode and transmit normative commitments, and variation in its use is variation in the normative signal a reader receives.

Large language models have rapidly become major producers of exactly this kind of language. LLMs now draft emails, advice, explanations, educational feedback, medical information, and policy text, and much of what they produce is directive, telling readers what they *should* consider, *need to* do, *cannot* do, or *have to* avoid. There is also growing evidence that the linguistic habits of these systems do not stay confined to machine output. Yakura et al. (2024) document that words statistically favored by ChatGPT have measurably increased in human spoken communication since the model’s release. Set alongside the experimental evidence, reviewed above, that norm perception and compliance shift with the modal formulation of a message (Kuang & Bicchieri, 2024), this finding raises a question that is no longer hypothetical. If humans increasingly read, and perhaps increasingly echo, text generated by AI, could the way AI marks obligation come to shape the way humans understand what is obligatory?

The present study addresses the precondition of that question rather than the question itself. Before asking whether AI deontic language could reshape human normative understanding, we must establish whether AI deontic language actually differs from human deontic language, and if so, in what way. If LLMs reproduce contemporary human patterns of modal usage, the influence question is moot, because exposure to AI text would simply reinforce existing norms. If LLMs mark obligation differently, however, the character of the difference matters. A model that uses obligation language more sparingly, or that substitutes formal and procedural constructions for immediate and interpersonal ones, delivers a different normative signal even when the semantic content of its advice is identical.

Our answer, developed across three primary corpora, an external benchmark, two controlled replications, and a naturalistic eleven-model replication, is that the difference is real, consistent, and patterned in an initially surprising way. AI-generated text underuses positive deontic modals relative to contemporary human writers in every comparison we examine. The gap is not spread evenly across the modal system. It concentrates in the constructions through which contemporary writers mark immediate, personally inflected obligation, forms such as *should*, *have to*, *had to*, and the conversational negation *can’t*, while AI matches or exceeds human rates on constructions at home in formal and procedural prose, among them *must*, *need to*, and *cannot*. This asymmetry of register and stance, rather than any overall poverty of normative language, is the paper's core finding, and we state it here because the distinction between what a modal asserts and the interpersonal footing it claims organizes everything that follows.

That such a difference might exist is consistent with a growing literature showing that LLM-generated text is systematically distinguishable from human writing. Instruction-tuned models exhibit a distinctive grammatical and rhetorical profile, with denser nominalizations, more participial clauses, and characteristic patterns of passive voice (Reinhart et al., 2025), and parallel differences appear in news text (Liang et al., 2024). Yet within this literature, and consistent with Georgiou's (2026) rapid review of the field, modal verbs remain underexamined as a specific discourse-pragmatic marker of the difference between AI and human writing.

Closest to our study, Yao and Liu (2025) compared GPT-generated and human-authored academic book reviews in the *Journal of Pragmatics*. They found that ChatGPT used interactional metadiscourse markers more frequently overall than human reviewers, a pattern driven by the overuse of attitude markers, but significantly underused hedges and self-mention. Their analysis treats modal forms such as *might* and *must* as part of broader stance-marking systems, suggesting that AI-generated writing may reproduce some surface features of evaluative discourse while failing to emulate the balance of caution, conviction, and authorial presence characteristic of human writers.

Relatedly, Tantucci and Culpeper (2026) show that ChatGPT's responses to escalating human disputes are shaped by a tension between human-like conversational adaptation and alignment-driven restraint. Their account of an "AI moral dilemma" is relevant here. Both their paper and ours examine how LLMs reproduce some pragmatic features of human interaction while diverging in others.

A second question concerns historical and register calibration. The Google Books Ngram corpus allows modal frequencies in contemporary AI and human text to be compared with more than a century of published English prose. We use this comparison to ask where AI modal density falls relative to a long-run formal-prose baseline. If LLMs are shaped by books, documentation, web pages, and other structured written material, LLM modal usage may reflect the typical characteristics of formal written text rather than the more immediate and interpersonal deontic style of contemporary informal writing.

We examine positive deontic modals (*must*, *ought to*, *need to*, *has to*, *have to*, *had to*, *should*), negative deontic modals (*mustn't*, *must not*, *can't*, *cannot*, *couldn't*, *shouldn't*, *should not*, and related forms), and impossibility expressions (*impossible*, *not possible*, *unviable*, *impractical*) across three primary corpora, an out-of-sample benchmark, controlled prompt-matched replications on student essays and on a second model family (Claude), and a naturalistic eleven-model replication. We ask which constructions drive AI-human differences and what those differences imply about the pragmatic register of AI-generated obligation.

### 1.1. Deontic Modality as Pragmatic Stance

The modal constructions examined here differ in more than semantic strength. They also differ in the speaker stance and interactional footing they make available (Biber et al., 2021). *Should* commonly formulates recommendation or morally inflected advice. *Have to* and *had to* often locate obligation in practical constraints. *Must* can mark strong obligation, but in written registers it often carries a more formal or institutional tone, while the semi-modals *have to* and *need to* have risen markedly in recent decades (Leech et al., 2009). *Need to* frequently appears in instructional texts, where necessity is framed as task-oriented rather than morally compelled. Negative forms similarly differ: *can't* often reads as conversational and immediate, whereas *cannot* is more typical of formal, institutional, or technical prose.

These distinctions matter for corpus pragmatics because a frequency difference across forms is also a difference in the distribution of pragmatic resources. A text that substitutes *need to* or *cannot* for *should*, *have to*, or *can't* may preserve a broad sense of necessity while changing the interpersonal stance. For this reason, we treat modal frequencies as evidence about register and stance and not merely as counts of interchangeable grammatical items.

### 1.2. Large Language Models, Register, and Pragmatic Style

The linguistic style of LLMs is determined in large part by training data and fine-tuning procedures. Models trained primarily on published books, academic papers, and structured web content will absorb the modal norms of those registers. Models fine-tuned for instruction-following will further adjust their output toward the stylistic norms of the feedback signal. The training corpora of early foundational models such as GPT-3 drew on Common Crawl (large-scale web scrapes), digitized books, Wikipedia, and curated link-filtered content (Brown et al., 2020). The corpus composition of later frontier models is largely undisclosed regardless of developer (OpenAI, 2023, 2025), but the recipes that are documented converge on the same predominantly web- and book-based mixture. The disclosed Llama pretraining data consists mostly of web scrapes supplemented by books, encyclopedic content, and code (Touvron et al., 2023), openly documented corpora built for model pretraining share this composition (Gao et al., 2020; Soldaini et al., 2024), and Google describes Gemini's pretraining data as web documents, books, and code (Gemini Team, 2023). Although this mixture is contemporary by volume, and consequently dominated by web text rather than historical documents, it skews heavily toward formal, edited prose and substantially underrepresents the informal spoken and written registers of everyday life (Bender et al., 2021).

For modal language specifically, this pragmatic shaping could take two distinct forms. First, LLMs might use modals at an overall rate different from that of contemporary unpublished speech, reflecting a training corpus dominated by formal registers in which strong deontic marking (*must*, *have to*, *had to*) could be more or less common than in informal speech and writing. Second, models might show asymmetric deviations across specific constructions, matching human rates on formal deontic modals (*must*, *cannot*) while underusing informal or colloquial forms (*have to*, *had to*, *can't*). The register asymmetry account predicts the latter pattern; that is, AI should approximate the modal resources of the formal written prose on which it was trained, but depart from the contemporary patterns of expressing obligation and necessity that are not fully incorporated in the training materials.

Our focus in this paper extends the metadiscourse approach taken by Yao and Liu (2025) and the broader discourse/pragmatic cue family identified by Georgiou (2026). While the existing work has examined broad categories of text and language, the present study specifically examines modal constructions as pragmatic resources for marking obligation and interpersonal force.

---

## 2. Method

### 2.1. Datasets

We drew on three corpora that pair AI-generated and human-authored text across distinct contemporary genres: the gsingh narrative dataset, the HC3 question-answer corpus, and a WritingPrompts fiction corpus we constructed for this study.

**gsingh narrative dataset.** This dataset (gsingh1-py/train, available via HuggingFace) pairs human-authored short narratives sourced from New York Times articles with GPT-4o-generated versions produced from the same prompts. The dataset contains 4,214,116 words of AI-generated text and 5,574,576 words of human-authored text.

**Hello-SimpleAI Human ChatGPT Comparison Corpus (HC3).** The HC3 corpus (Guo et al., 2023) is a large-scale parallel corpus containing question-answer pairs from human experts and from ChatGPT (GPT-3.5/4) across open-domain, financial, medical, legal, and psychological domains. The English-language portion of the corpus, which we analyze here, contains 9,517,180 words of ChatGPT-generated text and 13,969,622 words of human-authored text.

**WritingPrompts fiction dataset.** This dataset (Fan et al., 2018; available via HuggingFace as euclaise/writingprompts) consists of short fiction stories written by contemporary human authors in response to creative writing prompts posted on Reddit's r/WritingPrompts community. To construct a parallel AI corpus, we randomly sampled 150 prompts (random seed = 42) from WritingPrompts stories of 150–800 words, retaining only prompts with at least 20 characters after stripping community tags (e.g., [WP], [EU]). Each prompt was submitted to GPT-4o (model: gpt-4o; generated 2026-04-13) with a system prompt instructing the model to write as a creative fiction writer with natural narrative voice and not to summarize or explain, and a user prompt requesting a short story of approximately 400 words. Generation parameters were temperature = 0.9 and max_tokens = 600. Stories shorter than 100 characters were excluded (0 of 150); no other filtering or editing was applied to AI outputs. The resulting corpus contains 67,202 words of AI-generated fiction and 65,709 words of contemporary human-authored fiction.

**Cross-model generation (Claude).** To test whether the deontic gap generalizes beyond a single model family, we generated a parallel set of 150 stories from Anthropic's Claude (model: claude-sonnet-4-6; generated 2026-05-30) using the identical 150 prompts, system prompt, user prompt, and generation parameters (temperature = 0.9, max_tokens = 600) used for the GPT-4o corpus. The only difference between the two AI corpora is the model itself; holding genre, prompt, and decoding conditions constant isolates model family as the sole source of variation. The resulting Claude corpus contains 59,899 words.

**ELEPHANT dataset.** Cheng et al. (2026) assembled a benchmark to measure social sycophancy across eleven large language models. It pairs 3,027 open-ended personal-advice queries, each with a crowdsourced human response, and 2,000 posts from the subreddit *r/AmItheAsshole*, each with the top-rated human comment, with responses to the same prompts from eleven independently developed models (GPT-4o, GPT-5, Claude, Gemini, DeepSeek, Qwen, and Llama and Mistral variants). Because these responses were generated by the original authors rather than under our decoding conditions, the dataset affords a naturalistic multi-model comparison rather than a controlled one. Applying a 50-word eligibility threshold to each response, the analytic sample comprised 2,692 human responses (334,802 words) and 55,070 model responses (18,978,681 words, a mean of roughly 1.7 million per model). The threshold retained almost all model responses (99.6%) but only 54% of the shorter human responses, so the eligible human sample is weighted toward the longer responses in each source.

**Pangram benchmark.** We analyzed the publicly released benchmark corpus from Pangram Labs, a commercial AI-text detection company, released with the technical report describing their detection system (Emi & Spero, 2024). The dataset contains 1,976 documents labeled as human-authored or AI-generated and spans ten domains (creative writing, student writing, scientific writing, books, encyclopedic text, news, email, reviews, blog posts, and short-form question-and-answer). The AI documents are drawn from eight different large language models, including GPT-3.5, GPT-4, Google Gemini Pro, Mistral 7B, and LLaMA 2. After excluding documents shorter than 50 words, the analytic sample comprised 1,925 documents (997 human, 928 AI).

**ASAP-AES student-writing replication.** We also conducted a controlled prompt-matched replication on student persuasive essays drawn from the Hewlett Foundation Automated Student Assessment Prize (ASAP-AES; Hamner et al., 2012; obtained via the public HuggingFace mirror TasfiaS/ASAP-AES). We sampled 50 essays each from prompt sets 1 and 2 (persuasive essays on the effects of computers and on library censorship, written by 8th- and 10th-grade students) and prompt set 8 (narrative essays on the role of laughter, 10th grade), for 150 paired prompts in total (random seed = 42). For each sampled human essay, we generated a paired GPT-4o response (model: gpt-4o; generated 2026-05-23) to the same prompt with a system instruction to write as a student responding to the assigned prompt, temperature = 0.9, and max_tokens scaled to the median human length for each set. We excluded source-based prompts (sets 3–5), which require access to a specific reading passage, so that the AI model and the original student writers responded to the same prompts under comparable input conditions. The resulting corpus contains 66,781 words of AI-generated text and 67,545 words of student writing.

### 2.2. Historical Comparison: Google Books Ngram Corpus

To situate AI and human modal rates in historical context, we retrieved per-phrase frequency data from the Google Books Ngram Viewer (Michel et al., 2011) for the years 1920–2022, using the ngramr package (version 1.9.0) in R. We drew on two Google Books corpora. The general American English corpus (en-US) serves as the century-long published-prose baseline against which we era-match all of our datasets, and the English Fiction corpus (en-fiction), a genre-specific corpus of published fiction, provides a fiction-matched baseline for the WritingPrompts comparison shown in the fiction panel of Figure 3. For each corpus we retrieved raw count-based frequencies (smoothing = 0) and converted them to per-10,000-word rates by treating the Ngram proportions as per-word frequencies and multiplying by 10,000. This procedure places AI rates, contemporary human rates, and historical book frequencies on an identical scale, enabling direct era-matching.

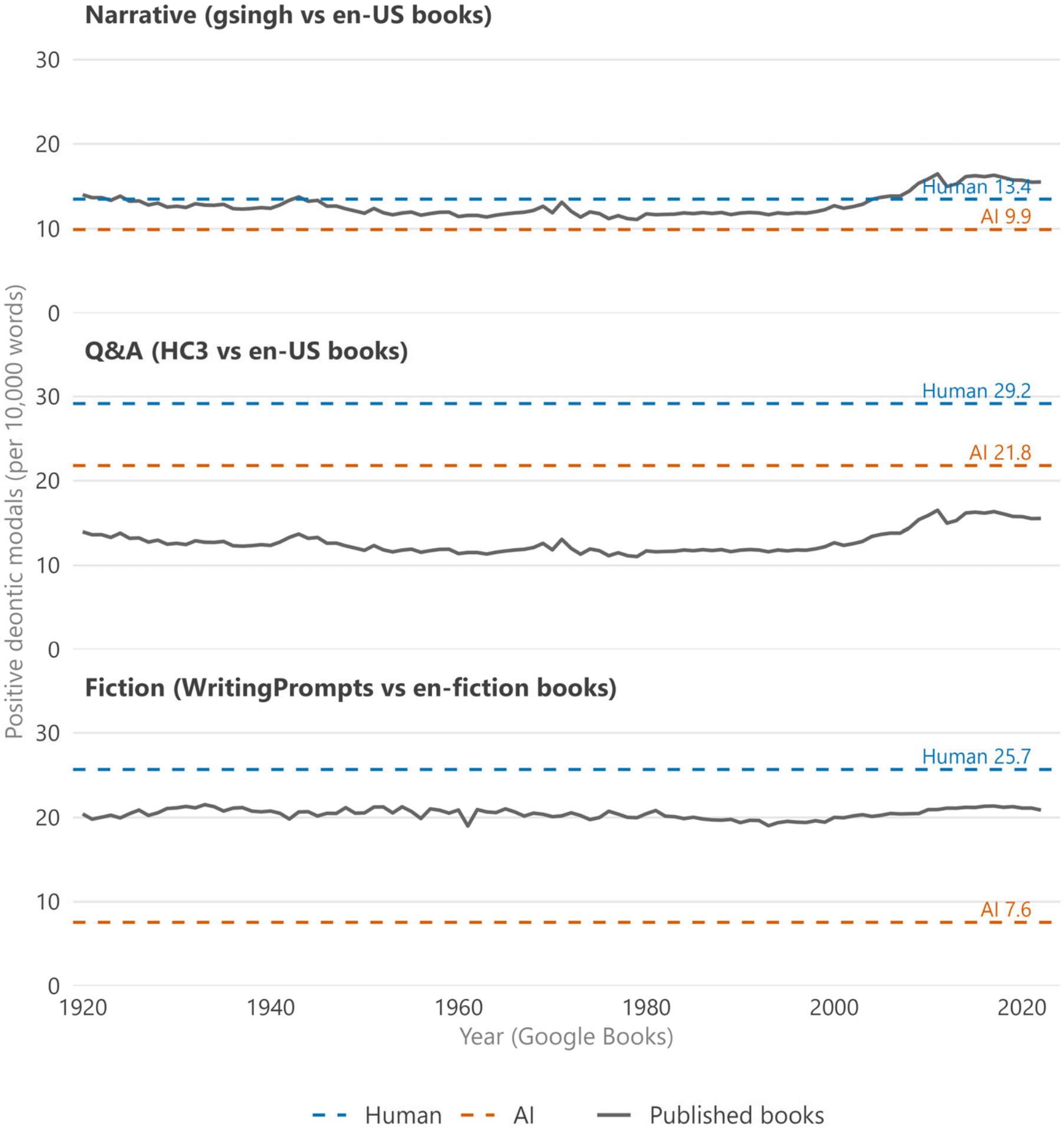

*Figure 3.* Historical trajectory of positive deontic-modal frequency in Google Books (grey line, per 10,000 words), with contemporary human and AI corpus rates (dashed lines) overlaid for three genre-matched comparisons: narrative (gsingh) and question-and-answer (HC3) against the en-US book series, and fiction (WritingPrompts) against the en-fiction series (a separate Google Books corpus restricted to published fiction). Within each panel, the grey line is the century-long Google Books trajectory and the two dashed lines mark the contemporary human and AI corpus rates. In every panel the AI rate falls within or below the century-long book range, whereas the human rate meets or exceeds its upper edge, most clearly in fiction.

For era matching, we identified the year in each Ngram series at which the historical frequency most closely approximated the observed AI or human rate, using minimum absolute deviation as the criterion.

### 2.3. Modal Phrase Lists

We operationalized positive deontic modals as the set {*must*, *ought to*, *need to*, *has to*, *have to*, *had to*, *should*}; negative deontic modals as {*mustn't*, *must not*, *oughtn't to*, *can't*, *cannot*, *couldn't*, *mayn't*, *mightn't*, *hasn't to*, *hadn't to*, *shouldn't*, *should not*}; and impossibility terms as {*impossible*, *not possible*, *unviable*, *impractical*}. Phrase detection used whole-word boundary matching (regex \b) with case-insensitive matching applied after lowercasing. Rates are expressed as occurrences per 10,000 words.

These phrase sets operationalize targeted constructions with a clear theoretical rationale rather than exhausting the English modal system. Several analytic choices deserve acknowledgment. First, impossibility terms differ from modal auxiliaries in grammatical form and function and are treated throughout as a secondary, exploratory comparison. Second, a central limitation of our methods is that the same form can perform different functions in context. We therefore interpret the corpus counts as measuring the distribution of modal *resources* rather than assigning a deontic interpretation to every token. The phrase-level decomposition below is designed to connect these aggregate patterns to pragmatic contrasts among constructions, such as the difference between conversational immediacy (*can't*, *have to*) and more formal or procedural necessity (*cannot*, *need to*).

### 2.4. Analytic Approach

All analyses were conducted in R. The unit of analysis is the corpus: all text within each source–dataset combination was pooled, and rates are expressed as counts per 10,000 words. For phrase-level analyses, we computed the gap between AI and human rates for each construction within each dataset, then averaged these gaps across datasets for the correlation analysis. Historical change for each modal construction was computed as the difference in Google Books (en-US) frequency between 1960 and 2019. Pearson correlations were computed between historical change and the AI–human modal gap across constructions, separately for positive ($k$ = 7) and negative ($k$ = 9) modals. For the negative modals, contracted and uncontracted variants of the same construction (e.g., *can't* and *cannot*; *shouldn't* and *should not*) were aggregated onto the single Ngram canonical form (e.g., *can not*) before the correlation, because the Google Books Ngram series records only the uncontracted variants. Given the small number of constructions, these correlations should be interpreted as descriptive summaries of the direction and consistency of the pattern rather than as precise parameter estimates, and we do not report inferential tests for them.

To assess whether the AI–human difference holds below the pooled-corpus level, we conducted document-level robustness analyses on both corpora in our study with prompt-matched paired designs: WritingPrompts (150 paired stories) and ASAP-AES (150 paired essays). For each paired prompt we computed positive-modal rates per document (counting only documents of at least 50 words). We then (a) bootstrapped the human–AI gap by paired resampling of prompts ($B$ = 2,000), (b) tested the paired difference with a Wilcoxon signed-rank test, and (c) fit a Poisson generalized linear mixed model with positive-modal count as outcome, source (human vs. GPT-4o) as fixed effect, log word count as offset, and prompt as random intercept. We report the WritingPrompts and ASAP-AES robustness checks here because their prompt-matched pairing yields the most defensible within-document contrasts.

For the independent out-of-sample replication on the Pangram benchmark, we computed positive-modal rate per document and assessed the direction and magnitude of the human–AI difference using a Wilcoxon rank-sum test, both overall and within each Pangram-defined domain. As a supplementary check on whether the signal separates individual documents rather than only populations, we also computed the area under the ROC curve for positive-modal rate as a classifier of the human/AI label (reported in the supplementary material).
Finally, to test whether the positive-modal underuse is specific to deontic constructions rather than a byproduct of the many dimensions on which AI and human word frequencies differ, we computed the GPT-4o-to-human frequency ratio for the 300 most frequent words across the three corpora with paired human and GPT-4o text (WritingPrompts, ASAP-AES, and the ELEPHANT advice questions) and located the positive deontic modals within that reference distribution. The deontic modals fell at the 3.7th percentile, more underused by GPT-4o than roughly 96% of common words, confirming that the pattern is specific and extreme rather than a generic difference in word frequency; we report this specificity check alongside the multi-model comparison in the Results.

## 3. Results

### 3.1. Overall Differences in the Pragmatic Marking of Obligation

Across all corpora, large language models used positive modals (e.g., *must*, *should*, *have to*) at substantially lower rates than human authors. Table 1 shows that across the three primary corpora and the ASAP-AES controlled replication, the AI rates (9.9, 21.8, 7.6, and 25.3 positive modals per 10,000 words for gsingh, HC3, WritingPrompts, and ASAP-AES, respectively) fell below human rates (13.5, 29.2, 25.7, and 52.9 per 10,000 words) in every comparison. AI's positive-modal shortfall was largest in the ASAP-AES student-essay corpus, where the gap of 27.5 per 10,000 words between high school student writers and GPT-4o represents the starkest register contrast in the dataset; the WritingPrompts fiction gap (18.1) was the second largest. The same direction held for negative modals: human authors used negative deontic forms at higher rates than AI in all four corpora (4.7 vs. 2.0 in gsingh, 10.0 vs. 3.5 in HC3, 13.4 vs. 8.0 in WritingPrompts, and 18.7 vs. 7.3 in ASAP-AES), with the gap concentrated in contracted forms (*can't*, *shouldn't*) and the formal uncontracted variants (*cannot*, *should not*) modestly favoring AI. Impossibility terms were used at low rates by both AI and humans and showed no consistent pattern across datasets.

*Table 1*
*Modal Rates (per 10,000 words) Across Primary Corpora and Controlled Replication Samples*

| Dataset | Genre | AI pos. | Human pos. | AI neg. | Human neg. |
|---|---|---|---|---|---|
| gsingh | Narrative | 9.90 | 13.45 | 2.02 | 4.73 |
| HC3 | Q&A | 21.80 | 29.16 | 3.52 | 9.98 |
| WritingPrompts | Fiction | 7.59 | 25.72 | 8.04 | 13.39 |
| ASAP-AES | Student essays | 25.31 | 52.85 | 7.34 | 18.65 |
| *M* | — | 16.15 | 30.30 | 5.23 | 11.69 |

*Note.* Rates are counts per 10,000 words. pos. = positive deontic modals; neg. = negative deontic modals. AI sources are GPT-4o (gsingh, WritingPrompts, ASAP-AES) and ChatGPT (HC3). Human sources are original paired texts within each corpus. Genres: gsingh = narrative non-fiction; HC3 = question-and-answer/informational; WritingPrompts = contemporary fiction; ASAP-AES = student persuasive and narrative essays. Impossibility terms, treated throughout as a secondary comparison, occurred at low rates (AI/human

per 10,000 words): gsingh 0.10/0.44; HC3 1.39/1.25; WritingPrompts 1.93/0.91; ASAP-AES 2.10/0.89 ($M$ = 1.38/0.87).

---

### 3.2. Robustness of the Paired-Design Gaps Below the Corpus Level

A natural concern about pooled corpus rates is that aggregate differences may be carried by a small number of documents or concentrated in a subset of prompts. Our two prompt-matched corpora, ASAP-AES (student persuasive and narrative essays) and WritingPrompts (Reddit creative fiction), allow direct within-document checks.

In ASAP-AES, across 149 paired essays meeting the 50-word threshold, the mean per-document positive-modal rate was 65.0 per 10,000 words for human writers (median = 34.9, $IQR$ = [0, 90.9]) versus 30.7 per 10,000 words for GPT-4o (median = 16.4, $IQR$ = [0, 55.7]). On both sides the mean sits well above the median, reflecting right-skewed distributions in which the gap is carried by a subset of modal-dense documents rather than by a uniform per-document shift, a point we return to below. The mean paired difference (human minus AI) was 34.1 per 10,000 words. A paired bootstrap over prompts ($B$ = 2,000) produced a 95% confidence interval of [19.0, 36.8] per 10,000 words for the corpus-level (word-weighted) gap of 27.5 reported in Table 1. A paired Wilcoxon signed-rank test confirmed the asymmetry ($V$ = 6,084, $p < .001$). A Poisson mixed model treating prompt as a random intercept and log word count as offset estimated the GPT-4o rate at 0.48 times the human rate (95% CI [0.40, 0.57]).

The WritingPrompts corpus reproduces the same pattern. Across 150 paired stories, the mean per-document positive-modal rate was 26.8 per 10,000 words for human writers (median = 18.0, $IQR$ = [0, 38.9]) versus 7.8 for GPT-4o (median = 0, $IQR$ = [0, 21.3]). The mean paired difference was 19.0 per 10,000 words, and a paired bootstrap produced a 95% CI of [13.6, 23.1] for the corpus-level (word-weighted) gap of 18.1 (paired Wilcoxon $V$ = 4,969, $p < .001$; Poisson GLMM rate ratio 0.30, 95% CI [0.22, 0.40]). The aggregate gaps in both corpora reported in Table 1 are therefore not artifacts of pooling: they survive prompt-matched, within-document comparison and a model that accounts for differences in document length and prompt-level variability. Figure 1 shows the WritingPrompts pairs directly. Most GPT-4o stories contain no positive deontic modal at all and sit on the left axis, whereas their human counterparts range widely up the vertical axis, so the pairs fall overwhelmingly on the human side of the equal-use line.

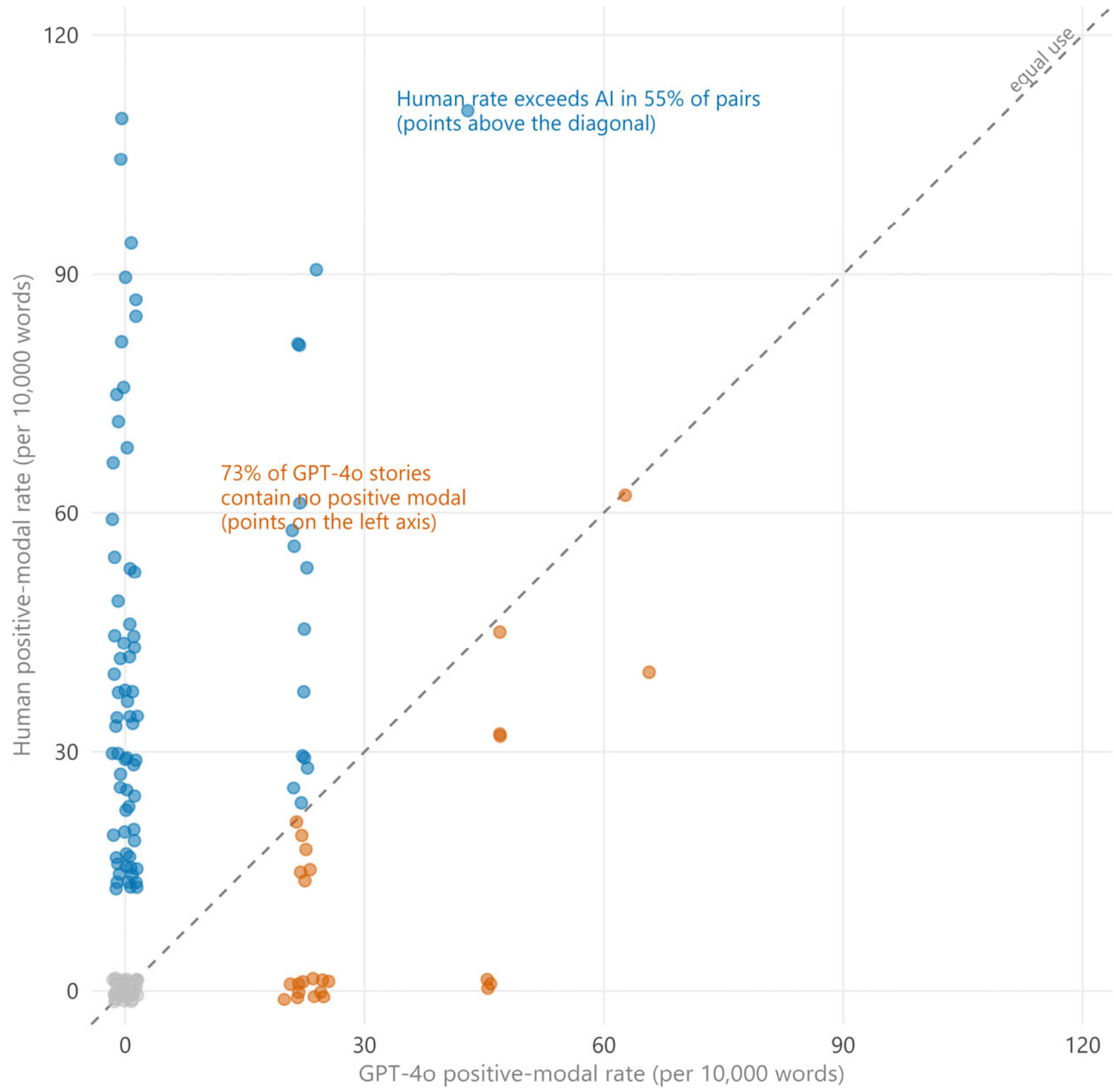

*Figure 1.* Per-document positive deontic-modal rates (per 10,000 words) in the WritingPrompts paired corpus, one point per prompt, comparing the human-authored story (vertical axis) with the GPT-4o story generated from the same prompt (horizontal axis). The dashed line marks equal use. Most GPT-4o stories contain no positive deontic modal (points on the left axis), whereas their human counterparts range widely, so the human rate exceeds the AI rate in the majority of pairs (points above the diagonal). Documents of at least 50 words; both axes are truncated at 120 per 10,000 words.

We note one nuance. Because many short documents on either side contain no positive modal at all (Figures 1 and S2), the gap is carried by the right tail of the human distribution rather than by a uniform per-document shift. Human rates exceed AI rates in 57.0% of ASAP-AES prompt pairs and 54.7% of WritingPrompts prompt pairs, are tied (often at zero) in many of the remainder, and exceed them substantially in the upper quartile. The register asymmetry the corpus rates describe is therefore best understood as a difference in how often, and how densely, human writers reach for interpersonal deontic resources, not a difference visible in every paired set of documents.

### 3.3. Out-of-Sample Replication Across Domains

To assess whether the modal-frequency pattern generalizes to independent data, and whether the difference is visible at the level of individual documents rather than only in the aggregate, we analyzed the publicly released Pangram benchmark (Emi & Spero, 2024; 1,925 documents at least 50 words long, 997 human and 928 AI, spanning ten domains and eight large language models). Two findings emerge.
First, the corpus-level pattern replicates. On the Pangram benchmark, human-authored documents averaged 23.6 positive modals per 10,000 words, compared to 14.1 for AI-generated documents (Wilcoxon rank-sum test: $W = 393{,}842$, $p < .001$). The direction we observe in our primary corpora and the ASAP-AES replication, with humans using more positive deontic modals than contemporary instruction-tuned LLMs, holds in all ten Pangram domains by mean rate and in nine of ten by document-level rank (the exception is encyclopedic text, where the human and AI distributions overlap), including each of the three domains that most closely match our analytic corpora (creative writing, question-and-answer text, and news; Figure 2). The pattern thus replicates across independent samples, different human sources, and a different set of AI models than the GPT-4o and ChatGPT outputs in our primary corpora.

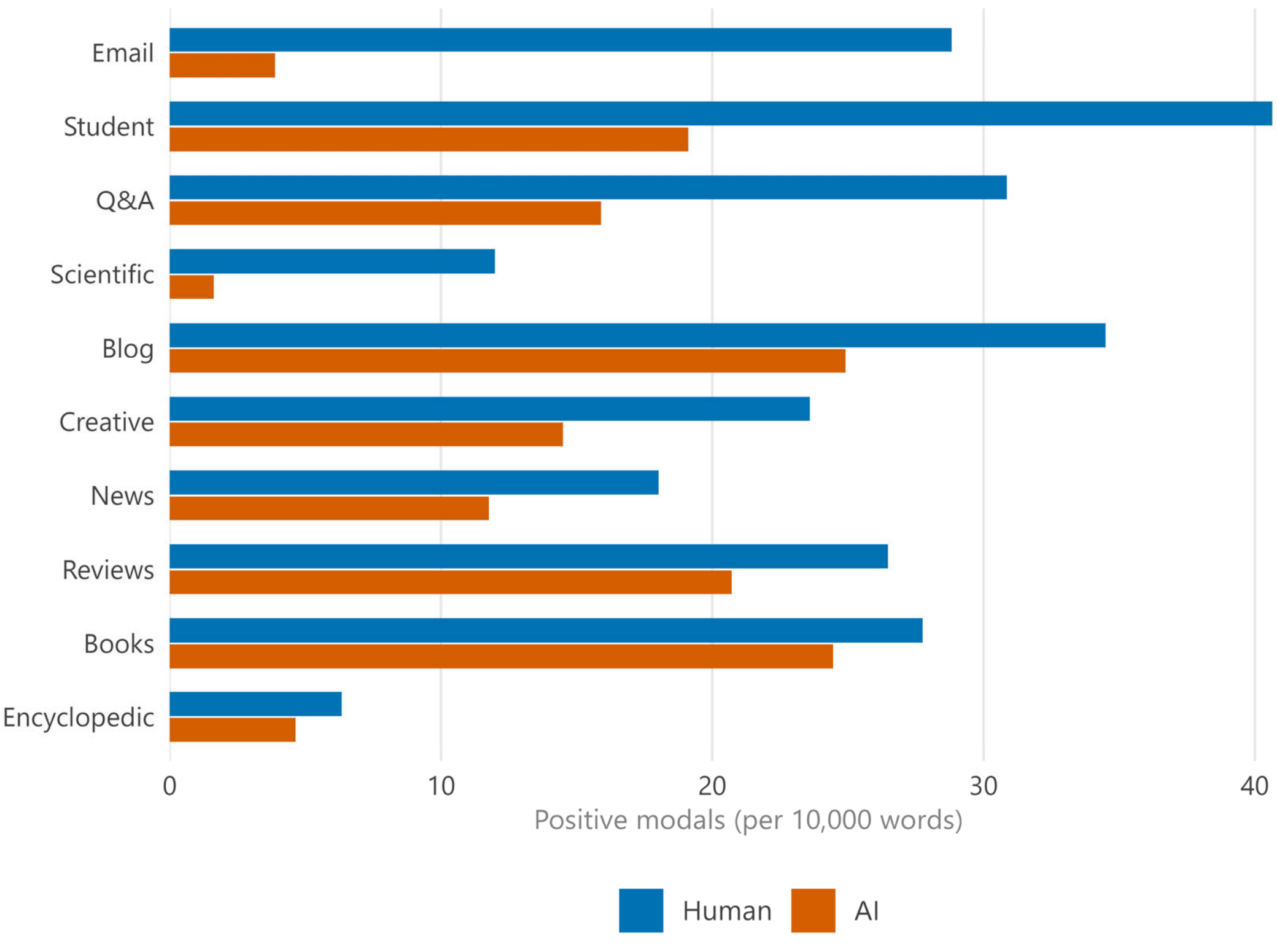

*Figure 2.* Positive deontic-modal rates (per 10,000 words) for human- and AI-authored documents across the ten domains of the Pangram benchmark (Emi & Spero, 2024). Human writers exceed AI in all ten domains by mean rate.
Second, the difference is a population-level pattern rather than a document-level marker. Although the direction is consistent across domains, positive-modal rate separates individual human and AI documents only weakly, so no single document can be reliably classified from its modal rate alone. What the replication establishes is that AI and human writers differ in the aggregate frequency with which they mark obligation, not that modal frequency functions as a marker of machine authorship in any one text. The per-domain discrimination statistics are reported in the supplementary material (Table S2).

### 3.4. Controlled Replication on Student Essays

The Pangram benchmark is useful for out-of-sample validation but is not a prompt-matched pragmatics corpus, and among its better-sampled domains the largest human–AI positive-modal gap appears in student writing (Figure 2 and Table S2; the email domain shows a numerically larger gap but rests on far fewer documents). To test that domain finding under a controlled prompt-matched design, we sampled 150 essay prompts from the Hewlett Foundation ASAP-AES corpus (50 each from two persuasive sets and one narrative set) and generated paired GPT-4o responses to the same prompts. The pattern is stronger than in any of our primary corpora: high school students used positive modals at 52.9 per 10,000 words, compared with 25.3 for GPT-4o. The paired bootstrap and Wilcoxon results are reported in the Robustness subsection above (95% CI for the corpus-level gap [19.0, 36.8] per 10,000 words; $V = 6{,}084$, $p < .001$; Poisson GLMM rate ratio 0.48 [0.40, 0.57]). Negative modals showed the same direction (human 18.7 vs. AI 7.3 per 10,000 words), with the gap again concentrated in contracted forms (*can't*, *couldn't*, *shouldn't*) while AI matched or slightly exceeded human rates on the formal uncontracted equivalents (*cannot*, *should not*), the same asymmetry observed in our three primary corpora.

One construction broke the pattern observed in our primary corpora. *Need to*, which AI overuses in instructional and question-answering text (gap of +2.3 per 10,000 words in HC3) and uses at near-human rates in narrative non-fiction, was substantially *underused* by AI in the persuasive-essay context (AI 0.3 vs. human 7.7 per 10,000 words). This genre-conditionality suggests that AI's reliance on *need to* is not a general preference but a register-specific feature of procedural and Q&A contexts: AI defaults to *need to* when the prompt invites task-oriented framing and to *should* and *must* when the prompt invites persuasion. We return to this finding in the Discussion.

### 3.5. Historical Calibration Against Published-Prose Register

To calibrate AI modal rates against the historical published-prose record, we compared AI and human rates to the Google Books Ngram corpus (en-US), converting Ngram proportions directly to per-10,000-word rates. This placed AI rates, human rates, and historical book frequencies on an identical scale, enabling era matching as a heuristic rather than a literal claim of temporal placement: the key question is not which decade AI "writes like," but where its modal density falls relative to the range of formal published-prose norms across the past century.

The WritingPrompts dataset provides the clearest era-matching contrast. Contemporary human fiction writers in this corpus used positive modals at 25.7 per 10,000 words, falling near the 2011 level in the en-US Ngram series, while GPT-4o completions of the same prompts averaged only 7.6 per 10,000 words, falling near the 1979 level. This gap of 18.1 per 10,000 words substantially exceeds what any decade-to-decade comparison in the en-US series shows, reflecting not an era difference but a register difference: contemporary informal fiction writers use deontic modals at frequencies that have outpaced the entire 20th-century published-book record. A parallel pattern emerged in the general corpus: AI-generated text from the gsingh dataset averaged 9.9 positive modals per 10,000 words (falling near the 1979 level in en-US), while ChatGPT text in HC3 averaged 21.8, near the 2011 level and close to the human HC3 rate. The narrower gap in HC3 likely reflects the informational, question-answering genre, in which the formal instructional register of LLM output is more congruent with human expert writing (Figure 3). Throughout, matched years should be read as approximate coordinates on the historical published-prose scale, not as claims that AI encodes the language of a specific decade; the Google Books series records the norms of edited, published text, which has always differed from informal human writing.

Contemporary human writers use modals at frequencies well above any historical book baseline, while AI language models produce text whose modal density falls within the range of published books (Figure 3). The magnitude of the AI–human modal gap thus tracks the distance between the genre of the comparison corpus and the formal written register of LLM training.

### 3.6. Phrase-Level Differences in Deontic Force and Stance

The aggregate gap between AI and human modal use masked meaningful differences across specific modal constructions. Decomposing positive modals by phrase revealed that the AI–human gap was concentrated in four constructions (*have to*, *should*, *had to*, and *has to*), while *need to* and *must* showed the opposite pattern, with AI matching or exceeding human rates (Figure 4). Averaged across the gsingh and HC3 datasets, humans used *have to* at 5.3 per 10,000 words compared to 1.7 for AI (gap: -3.6 per 10,000 words); *should* at 6.1 versus 4.6 (gap: -1.5); *had to* at 1.4 versus 0.3 (gap: -1.1); and *has to* at 1.5 versus 0.5 (gap: -1.0). By contrast, AI used *need to* at 5.3 versus 4.4 for humans (gap: +0.9), and *must* at 3.4 versus 2.5 for humans (gap: +0.9). The ASAP-AES replication, discussed above, shows that this *need to* pattern does not generalize to persuasive student writing, where AI sharply underuses the construction (0.3 vs. 7.7 per 10,000 words).

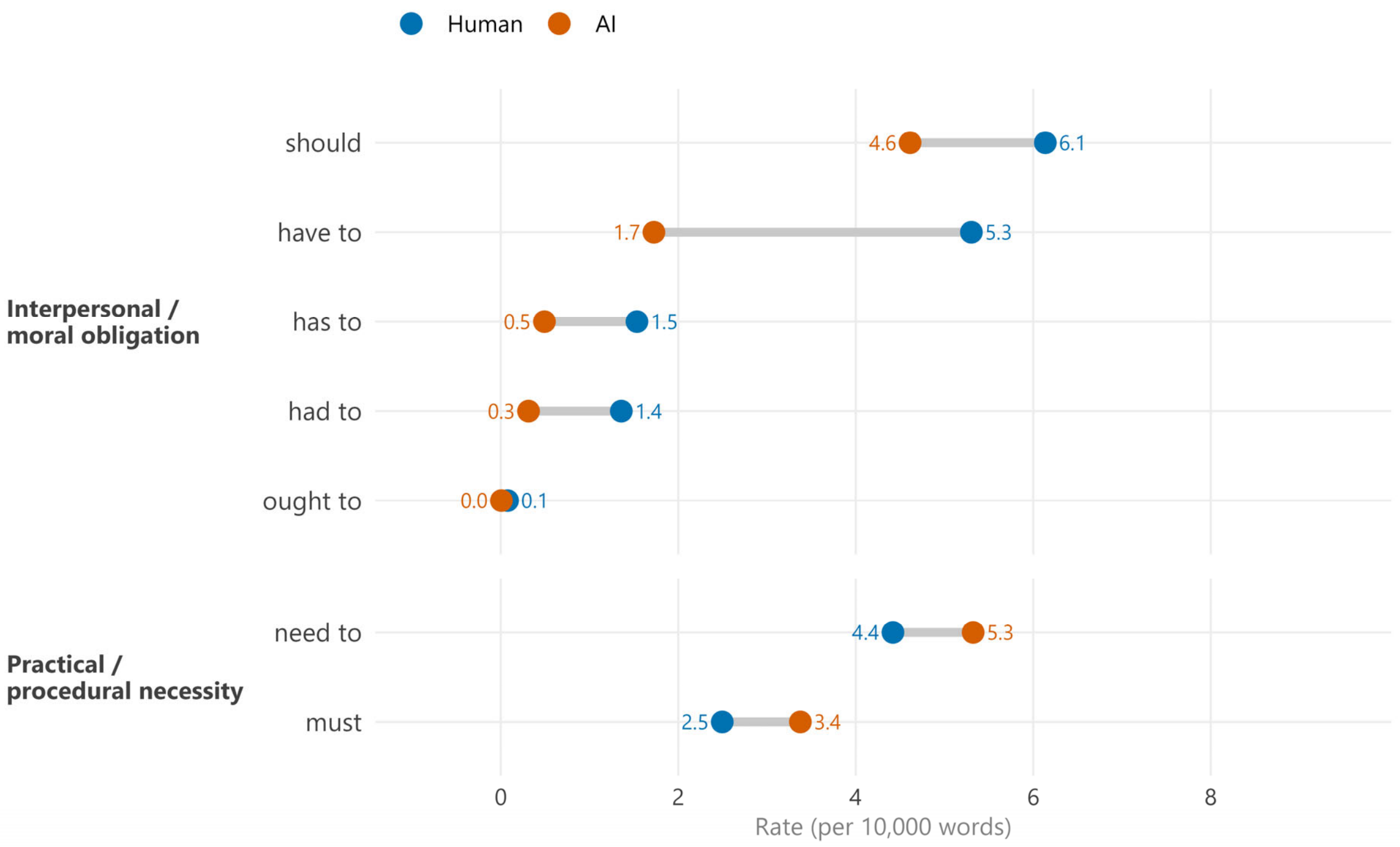

*Figure 4.* Phrase-level decomposition of positive deontic modals (per 10,000 words), averaged across the gsingh and HC3 corpora and grouped by pragmatic stance. Each row

connects the human rate (blue) and the AI rate (vermillion) for one construction. Human writers use the interpersonal-obligation modals (*should*, *have to*, *had to*, *has to*) more than AI, whereas for the practical-necessity modals *need to* and *must* AI matches or exceeds human rates.

Pragmatically, these findings suggest that in comparison to AI, human writers more often use forms associated with recommendation, personal constraint, and narrative compulsion (*have to*, *should*, *had to*). AI-generated text relies comparatively more on forms compatible with formal or procedural necessity (*must*, *need to*). This distinction is central to the register account: the AI–human gap is concentrated in the constructions most closely tied to immediate interpersonal obligation, not in modal language as such.

A parallel decomposition of negative modals showed an even sharper version of the same asymmetry. Averaged across the gsingh and HC3 datasets, the contracted forms drove a large human advantage, with *can't* at 4.9 vs. 0.9 per 10,000 words (gap: -4.0), *couldn't* at 0.9 vs. 0.1 (gap: -0.8), and *shouldn't* at 0.5 vs. 0.04 (gap: -0.5), while the formal uncontracted equivalents modestly favored AI: *cannot* (1.2 vs. 0.7; gap: +0.5) and *should not* (0.5 vs. 0.3; gap: +0.2). The contractions, which are characteristic of informal spoken and written registers, account for nearly the entire negative-modal gap; the uncontracted variants, which are typical of formal and institutional prose, are if anything overrepresented in AI output.

### 3.7. Illustrative Contrasts in Modal Stance

The phrase-level pattern can be seen in concrete prompt-matched contrasts. Consider one WritingPrompts pair in which both writers respond to a prompt about a mother who fears her daughter is becoming a "Disney princess" (singing, attracting animals, and receiving repeated proposals from princes). Both writers indicate that the situation cannot continue, but their use of modals conveys different emotional resonance. The human writer stages the conversation as panicked:

"Mary, you can't seriously be thinking—" "I *have* to! I *have* to, for my own safety!"

Across the human story, the mother's predicament is densely marked with interpersonal deontic forms: *should be proud*, *would need to*, *have to*, *can't*, *must be a witch*. Although GPT-4o also conveys doubt about the situation, its words reflect a more procedural, measured use of modal language:

"Dearest, we really *need to* talk about these proposals."

Both passages express necessity, but they represent necessity differently. The human writer reaches for constructions through which contemporary speakers mark lived compulsion (*have to*) and conversational impossibility (*can't*). The LLM renders the same scene through *need to*, a form more at home in procedural texts, and through the mental-state modal *couldn't* ("I couldn't help but smile"), which marks involuntary affect rather than obligation. The contrast is therefore not simply one of frequency but of footing: the human text frames obligation as urgent and personally compelled, a demand arising from the relationship itself, whereas the AI text frames it as impersonal, procedural management, the kind of necessity imposed by neutral rules rather than by what one person owes another.

### 3.8. Historical Change and the AI Modal Deficit

The phrase-level data allow a direct test of one specific version of the register account: if AI modal use simply reflects the modal norms of formal written text rather than contemporary informal usage, the AI–human gap should track the historical trajectory of each construction in the published-prose record. That is, constructions that have risen in books should be used at near-human rates by AI, while those that have declined should be underused. To assess this, we computed the change in Google Ngram frequency for each modal construction between 1960 and 2019 (en-US corpus) and correlated it with the AI–human gap across constructions (Figure 5). We treat the resulting correlations as descriptive, convergent evidence rather than confirmatory tests, because they are based on small numbers of constructions ($k$ = 7 positive, $k$ = 9 negative) and no inferential corrections are applied. The full 1920–2022 trajectory of each positive construction is shown in Figure S1, and several of them (notably *have to*, *had to*, and *has to*) rose only in recent decades rather than monotonically, so a single 1960-to-2019 difference is at best a coarse summary of their movement.

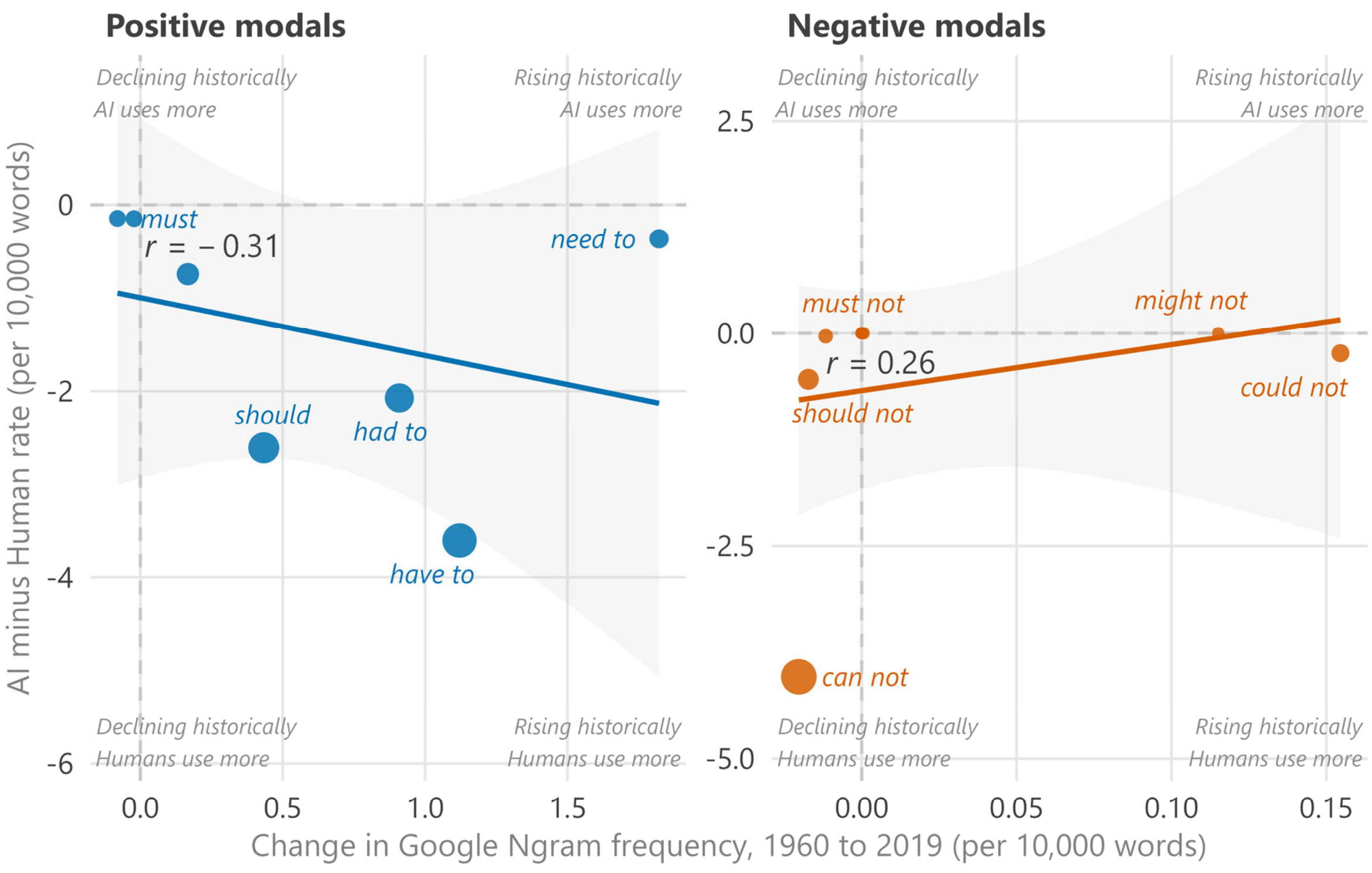

*Figure 5.* Relationship between historical change in Google Books (en-US) frequency from 1960 to 2019 and the AI-minus-human rate gap, shown separately for positive and negative modals. Point size is proportional to the absolute size of the gap.

For negative modals, historical change from 1960 to 2019 was positively correlated with the AI–human gap ($r$ = .26), in the direction the register account predicts: constructions that rose in books (e.g., *could not*, *might not*) showed near-zero gaps, while those that declined or remained stable in books (e.g., *can not* / *can't*, *should not* / *shouldn't*, *must not*) showed larger AI shortfalls. The correlation is modest, but the qualitative ordering of the negative-modal points is consistent with register fixation.

For positive modals, however, the correlation was weakly negative ($r$ = -.31), opposite the direction the simple register account predicts. *Need to* fits the register prediction (a 620% rise in Google Books since 1960, a rise independently documented in reference corpora by Leech et al., 2009, and the largest AI-favoring gap of any positive), and *must* fits weakly as well (roughly flat historical trajectory and a modest AI advantage). But the other five positives (*have to*, *had to*, *should*, *has to*, and *ought to*) are underused by AI regardless of whether they have risen or fallen in published books over the same period.
What is clear is that the AI–human gap is concentrated in the constructions through which contemporary humans mark immediate, personally inflected obligation, and that the register-fixation mechanism, while supported for negatives and for *need to* / *must* on the positive side, is not the whole story.

**3.9. Generality Across Model Families**

If the deontic gap reflects the formal-register character of LLM training data rather than a quirk of one developer's pipeline, it should appear in models built and tuned independently. We tested this directly by regenerating the 150 WritingPrompts stories with Anthropic's Claude (claude-sonnet-4-6) under conditions identical to the GPT-4o run. Claude used positive deontic modals at 9.7 per 10,000 words, far below the contemporary human rate of 25.7, and close to the GPT-4o rate of 7.6 (Figure 6). The mean paired difference between human writers and Claude was 17.1 per 10,000 words, with a paired bootstrap 95% CI of [10.9, 21.3] for the corpus-level (word-weighted) gap of 16.0 (paired Wilcoxon $V$ = 5,231, $p < .001$), and a Poisson mixed model with prompt as a random intercept estimated the Claude rate at 0.38 times the human rate (95% CI [0.28, 0.51]). The corresponding GPT-4o rate ratio in the same model was 0.30 (95% CI [0.22, 0.40]); the two confidence intervals overlap substantially, indicating that the two model families underuse positive deontic modals to a statistically indistinguishable degree. The shortfall fell on the same constructions in both models: *must*, *had to*, *should*, *have to*, and *need to* were all underused relative to human writers. That an independently developed model reproduces both the magnitude and the construction-level profile of the gap is difficult to reconcile with a pipeline-specific account; instead, it is consistent with what the register-fixation interpretation predicts: both families learn the modal norms of formal written text, and both carry those norms into creative writing.

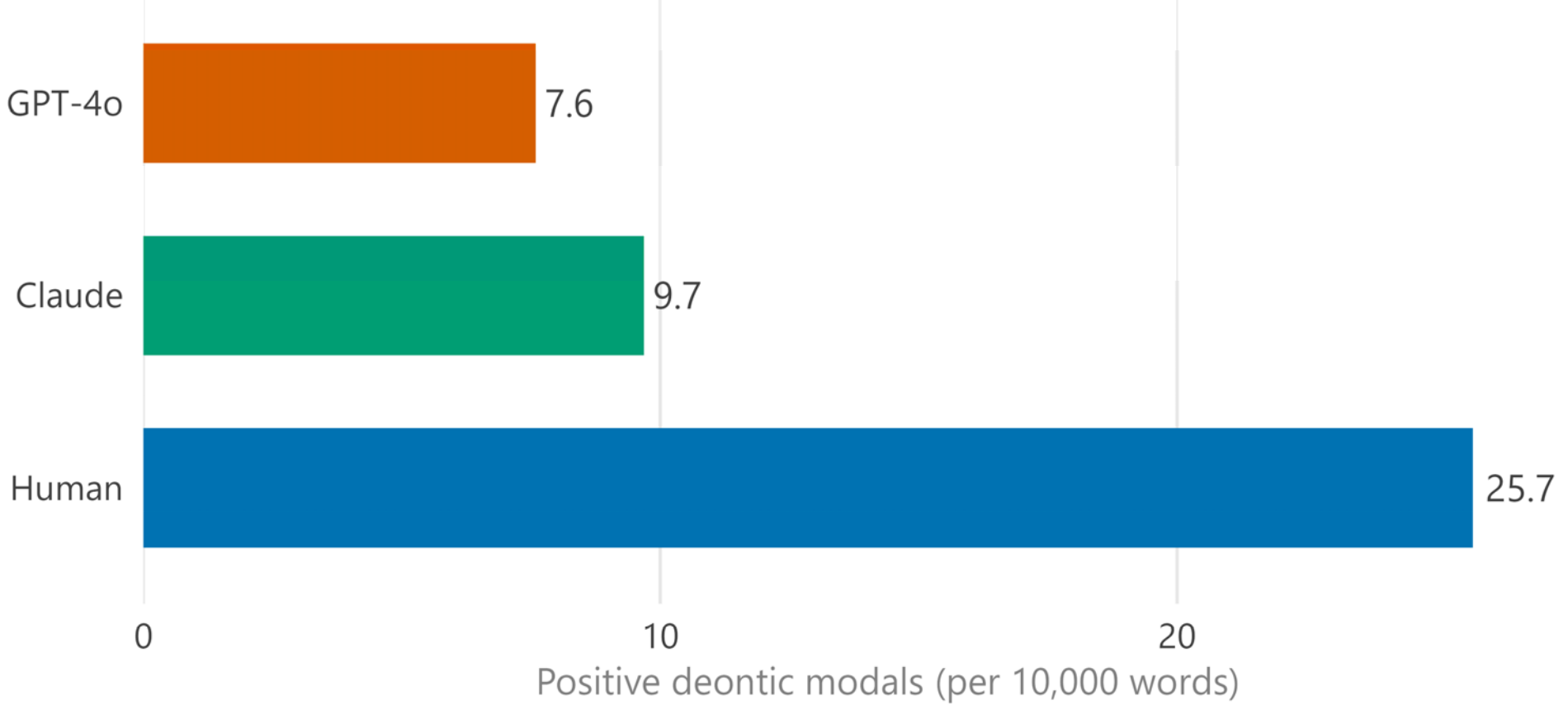

*Figure 6.* Positive deontic-modal rates (per 10,000 words) in the WritingPrompts corpus for contemporary human writers, GPT-4o, and Claude (claude-sonnet-4-6). All three sets of stories respond to the same 150 prompts under identical generation conditions; only the model differs between the two AI columns. Both model families fall far below the human rate.

A broader, naturalistic test of generality comes from the ELEPHANT corpus (Cheng et al., 2026), which pairs personal-advice and moral-judgment prompts with responses from human writers and from eleven separately developed models spanning seven organizations. Because these responses were generated by their original authors rather than under our decoding conditions, they complement the controlled Claude replication above. The pattern is uniform. Human advice-givers used positive deontic modals at 47.9 and 51.0 per 10,000 words across the two settings, near the top of the human range in this study, while every one of the eleven models fell well below (13.0 to 33.5; length-adjusted rate ratios of 0.23 to 0.66; Table S1 and Figure 7A). That eleven independently trained and aligned models reproduce the gap is difficult to reconcile with any developer-specific explanation.

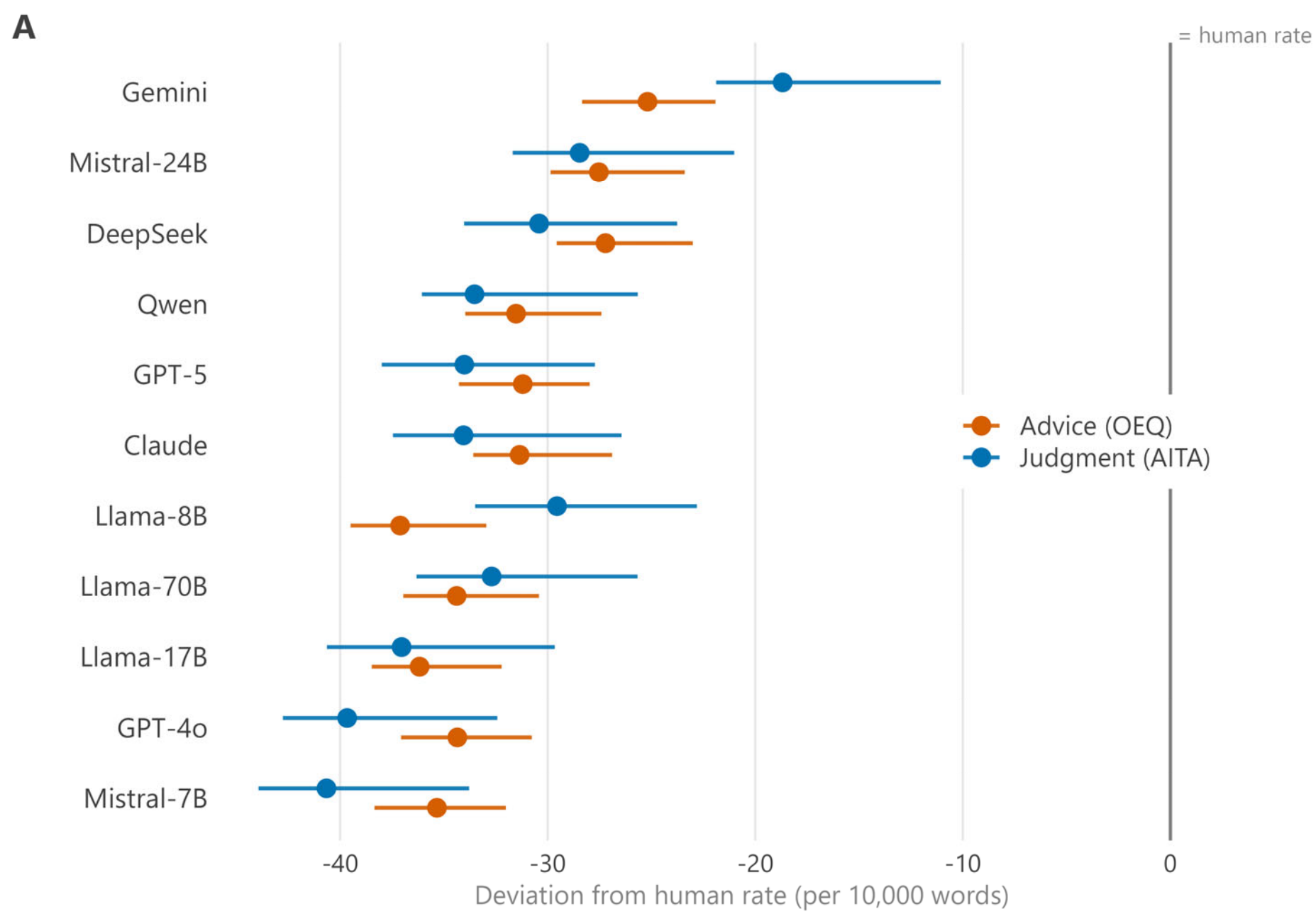

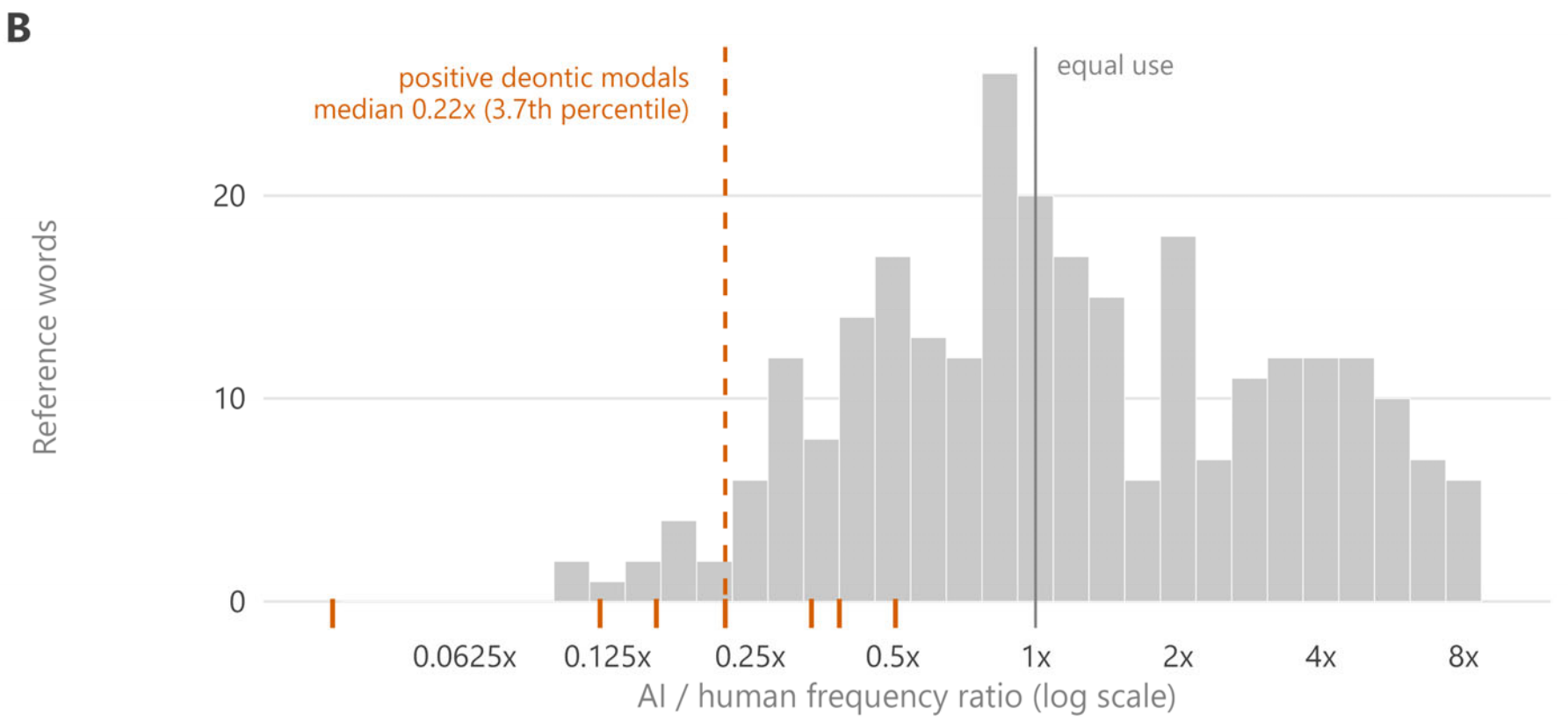

*Figure 7.* Generality and specificity of the deontic gap. (A) Mean paired gap from the human writer (model minus human, per 10,000 words) for each of the eleven ELEPHANT models (Cheng et al., 2026), with bootstrap 95% confidence intervals, in the advice (OEQ) and moral-judgment (AITA) settings; every model falls to the left of the human reference in both settings. (B) Distribution of GPT-4o-to-human frequency ratios (log scale) for the 300 most frequent words in the prompt-matched corpora (grey); the seven positive deontic modals (vermillion ticks) and their median (dashed line) fall in the far-left underuse tail, at the 3.7th percentile, indicating that the underuse is specific to the deontic constructions rather than a generic word-frequency difference.

The gap is also specific to the deontic constructions rather than a byproduct of the many ways in which AI and human word frequencies differ. Across a reference distribution built from the 300 most frequent words in the prompt-matched corpora, the seven positive deontic modals fall at the 3.7th percentile of the GPT-4o-to-human frequency-ratio distribution, more underused than roughly 96% of common words (Figure 7B). The deontic gap is therefore neither a single developer's artifact nor a generic property of how models render common vocabulary, but a specific and repeatable feature of how they mark obligation.

---

## 4. Discussion

We began with a simple question: do large language models use the language of obligation the way contemporary humans do? The answer, across three primary corpora, an independent benchmark replication, controlled prompt-matched replications on student essays and on a second model family (Claude), a naturalistic replication across eleven models in the advice-giving register, and document-level robustness checks, is consistently no. The evidence is consistent with a register-asymmetry account according to which LLMs reproduce the ways of expressing necessity characteristic of their training data while underusing constructions that contemporary human writers deploy to mark immediate obligation, recommendation, and interpersonal constraint.

Because the pattern proves so consistent across corpora and models, it is worth stating plainly what the design already forecloses. The prompt-matched WritingPrompts and ASAP-AES comparisons pair a human and an AI document written to the identical prompt, holding topic and genre constant by construction, and the accompanying Poisson mixed models express modal counts per word through a log-length offset. Neither differences in document length nor differences in the genres sampled can therefore account for the human–AI gap. The genre variation that does remain, the widening of the gap in fiction and persuasion and its narrowing in informational text, is a substantive feature of the pattern rather than an artifact of comparing unlike documents.

These findings extend Yao and Liu's (2025) conclusion that AI stance is genre-sensitive but incomplete. Their analysis located this incompleteness in the balance among hedges, boosters, attitude markers, and self-mention in academic book reviews. The present study shows a parallel imbalance in which AI-generated text does not simply use less normative language; it redistributes deontic force away from informal, interpersonal forms such as *should*, *have to*, and *can't* and toward more formal or procedural resources such as *need to* and *cannot*.

Our findings also converge with those of Tantucci and Culpeper (2026), who found that in rounds of human-AI disputes, AI consistently responded in less impolite ways than did humans. In this study and in theirs, AI text is less likely to rely on forceful, immediate marking, producing instead more measured responses. Interestingly, the convergence is apparent even though the two papers rely on substantially different methods. Tantucci and Culpeper used fine-grained analysis of a small number of AI-human disputes of multiple rounds, while ours relied on frequency analysis of multiple corpora.

Although neither paper provides definitive evidence on the matter, Tantucci and Culpeper (2026) locate the origins of AI's reluctance to express impoliteness in the training mechanisms that LLMs undergo while in our view the training register is also important in shaping the morally-relevant text generated by AI. Our work finds that the differences we report are genre-dependent; *need to*, for example, is over-represented in AI-generated text relative to human writing in instructional contexts, but relatively rare in persuasive essays. We also find the same pattern across independently developed models, in a controlled comparison of GPT-4o and Claude and in a broader naturalistic comparison spanning eleven models that likely overlap substantially in the training register but have undergone independent alignment training. These are strikingly different accounts of how the asymmetry arises. One locates it in the deliberate interventions of developers during alignment, the other in the constitution of the training corpora themselves, and the two carry very different implications for AI's future normative influence, because they imply very different answers to how much of the gap could be changed by adjusting the training process rather than the underlying corpora on which models are trained. Both mechanisms, alignment and the constitution of the training register, are likely to matter, and identifying their relative contributions is a pressing question for future research.

### 4.1. Register Asymmetry and Pragmatic Force

The clearest illustration of the register asymmetry comes from the WritingPrompts dataset, where contemporary human fiction writers (Reddit users responding to creative prompts) use positive modals at 25.7 per 10,000 words, while GPT-4o responses to the same prompts average only 7.6. The robust claim from this comparison is that AI modal density falls within the range of formal published-book prose across the past century, while contemporary informal fiction writers use deontic modals at frequencies that exceed the entire 20th-century book record. The matched years we report are approximate coordinates on that scale, not claims about which decade an AI "writes like." The 18.1-per-10,000-word gap is therefore best read as a register difference between formal published prose and contemporary informal writing, not as a temporal displacement.

Our findings do not support the conclusion that AI modal verb usage resembles a particular year or decade of historical text. One reason for this conclusion is that the era-matching results are genre-dependent. In the HC3 informational corpus, AI modal rates closely approximate contemporary human rates, with both falling near the 2011 mark in the Google Books series. The fiction gap is largest precisely where one would expect it if the driving factor is register, not era: fiction is the domain where informal, emotionally engaged human writing diverges most dramatically from the norms of formal published text. The register asymmetry account, positing that AI modal language reflects the modal norms of formal written text, which have remained relatively stable, rather than the norms of informal human writing, which have continued to rise, accounts for both the large fiction gap and the smaller informational gap within a single framework.

A related but distinct alternative interpretation deserves direct consideration. It is possible that LLMs produce historically older-seeming modal language because their training corpora contain a substantial proportion of historical text through sources such as Project Gutenberg, which supplies public domain works largely predating 1928. On this *historical average* account, the temporal displacement of AI modal language would reflect a genuine averaging over different eras of English prose. The present pattern is less consistent with this account than with a register-based interpretation. Modern LLM training corpora are dominated by contemporary text. Common Crawl constitutes by far the largest source by token count in corpora such as that used to train GPT-3, with digitized books comprising a relatively small weighted fraction of the total (Brown et al., 2020). More critically, the historical average account makes a prediction our data directly falsify. If AI modal language reflects an averaging over eras in which *need to* was rare, AI should markedly underuse this construction, because *need to* was far less common in mid-20th-century books than today. However, just the opposite is true: AI uses *need to* at human rates or above. This anomaly is difficult to explain under a simple historical-average account but is readily interpretable under a register account: *need to* is overrepresented in instructional and procedural web text, which dominates modern training corpora regardless of its historical era. The temporal displacement we observe is therefore better understood as register fixation. AI has absorbed the formal written norms of its training data, and that register happens to be one whose modal densities have not kept pace with the continuing informalization of everyday human language. Because this fixation should follow from the formal written text common to the training of all large language models rather than from any single developer's choices, it predicts that the gap should recur across independently built model families, a prediction confirmed both by our controlled Claude replication, in which GPT-4o and Claude underuse positive deontic modals to a statistically indistinguishable degree, and by a broader comparison in the advice-giving register, where all eleven independently developed models fall well below human rates.

### 4.2. Where AI Matches or Exceeds Humans: *Must* and the Genre-Conditional *Need To*

The two positive modals for which AI equals or exceeds human rates in our primary corpora, *must* and *need to*, illuminate the mechanism by which training data composition shapes modal language, but in different ways. *Must* has held roughly flat in published books between 1960 and 2019, and AI uses it consistently more than humans across our primary corpora (3.4 vs. 2.5 per 10,000 words, averaged across gsingh and HC3). *Must* flourishes in formal, regulatory, and definitional discourse, which are the registers that dominate documentation, encyclopedic content, and technical writing typical of LLM training data. AI's slight overrepresentation of *must* appears to be a stable feature of its deontic profile across genres.

*Need to* is more revealing because its AI–human relationship is genre-conditional. In the question-and-answer register, where instructional framings predominate, AI uses *need to* substantially more than humans (HC3: 8.66 vs. 6.35 per 10,000 words); in narrative non-fiction it tracks human rates; but in persuasive student essays it is sharply *underused* by AI (ASAP-AES: 0.3 vs. 7.7 per 10,000 words). This pattern is difficult to explain as a general AI preference but is readily interpretable as a register-specific feature of training-data composition: AI reaches for *need to* in contexts where its training corpora (instructional text, procedural documentation, FAQs, and question-answering corpora) make it the default form for marking necessity, and reaches for *should* and *must* in contexts (persuasion, argument) where those forms are the default in the same corpora.

The broader implication is that the AI–human modal gap is not a flat property of LLM output but is shaped jointly by training-data composition and by the register the prompt invites. The gap should be largest in genres such as fiction, personal narrative, persuasive argument, and emotional correspondence, where informal and interpersonally inflected deontic resources predominate in human writing but where AI's training data is comparatively thin. It should be smallest, and may even reverse for particular constructions, in genres where formal instructional registers dominate. The ASAP-AES result, which is the largest human–AI positive-modal gap in our study, is consistent with this account. The genre of persuasive student writing is a register in which human writers rely heavily on explicit recommendation and obligation, and it is the register where the gap is widest.

### 4.3. Implications for Pragmatic Interpretation and Future Research

The psycholinguistic literature reviewed earlier establishes that modal language encodes normative commitment and shapes the inferences readers draw. The constructions in which AI shows the largest gaps, *should*, *have to*, *had to*, are precisely those that convey the strongest and most personally immediate senses of obligation. *Should* carries a distinctive normative recommendation; *have to* conveys constraint or compulsion; *had to* retroactively frames past behavior as obligatory or compelled. AI's relative avoidance of these forms, combined with its preference for *need to*, *must*, and formal equivalents like *cannot*, produces a deontic register that is more formal, more distanced, and arguably less urgently normative than the way contemporary humans actually write about obligation. This is the paper's core descriptive finding.

Park et al. (2025) have shown that modal expressions in user prompts systematically bias the normative judgments that language models themselves render, indicating that modal marking carries normative weight within these systems. If AI systems systematically avoid the deontic constructions that convey personal immediacy and strong normative force, one possibility is that they produce text that is perceived as less commanding, less urgent, or less personally directed even when the semantic content of the obligation is identical. In domains where AI-generated text is used to convey guidance, advice, or recommendation this modal attenuation raises the question of whether consequences extend beyond stylistic preference.

A central question raised by these findings, and one that the present study cannot directly answer, is whether modal-sparse AI text produces weaker normative uptake in readers. There are reasons to imagine that this could occur. Kuang and Bicchieri (2024) report that injunctive framing (*you should*) produces stronger normative uptake than descriptive framing, and a broader literature on moralized language documents that small lexical shifts can alter how strongly a message registers as obligatory (Brady et al., 2017; Capraro & Vanzo, 2019). The same logic may reach beyond how a message is read to how a relationship is conducted, since in interpersonal settings whether a request is framed as a moral demand or as a practical necessity can itself shape how readily it is accepted, so a systematic drift toward procedural framing in the text people increasingly rely on is not obviously without consequence. Whether these effects generalize to the kind of modal-frequency differences we observe between AI and human text is an important question for future experimental work.

### 4.4. Limitations

Several limitations of the present study deserve acknowledgment. First, our three primary corpora, the Pangram benchmark, and the ASAP-AES replication together span ten or more contemporary genres but do not exhaust the range of contexts in which AI and human language differ. Spoken corpora, instant messaging, and professional writing in legal or medical settings may show different patterns. Second, as noted in the Method section, modal phrase lists are targeted rather than exhaustive, and key ambiguities (deontic vs. epistemic uses; excluded periphrastic forms) are not fully resolved by corpus-scale aggregation. Third, era matching is a heuristic calibration against the published-prose record. The Google Books corpus is weighted toward edited text, and matched years should be interpreted as rough placement within the formal-prose range rather than precise equivalence. Fourth, the document-level robustness analyses we report are restricted to WritingPrompts and ASAP-AES, where prompt-matched pairing makes the within-document comparison most defensible. The Pangram benchmark provides independent out-of-sample replication of the population-level pattern but does not, on its own, substitute for paired within-corpus analyses of gsingh and HC3. Additionally, the ASAP-AES human writers are 8th- and 10th-grade students writing under timed standardized-test conditions, a register characterized by both age and task constraints. The size of the human–AI gap we observe there reflects the persuasive-essay register's natural reliance on deontic constructions and should not be read as a comparison between AI and adult professional writers. Fifth, the historical-change correlation analysis was conducted on a small number of constructions ($k$ = 7 positive, $k$ = 9 negative) without inferential corrections; conclusions about the relationship between historical trajectory and the AI–human modal gap should therefore be treated as descriptive, although the broader finding that AI underuses positive deontic modals rests on the full corpus comparisons rather than on this small set. Sixth, the Pangram benchmark was constructed and released by the same developers whose detection system it evaluates, and its human/AI labels are the output of that construction process. We use the benchmark here because it provides an independent out-of-sample test of the modal-frequency pattern on data we did not select, but we do not treat the >99% detector accuracy reported on it as an independently validated figure. Finally, our findings characterize a particular generation of instruction-tuned models. Our controlled generations come from GPT-4o and a single Claude model, and the AI documents in the Pangram benchmark, though more diverse, are drawn predominantly from models released between 2022 and 2024. Because these systems are evolving rapidly, and because alignment procedures in particular are a moving target, later models may attenuate, preserve, or even reverse the deontic gap we document; whether it narrows, persists, or widens across successive model generations is a question the present design cannot resolve, and an important direction for future work. That the gap nonetheless recurs across GPT-4o, ChatGPT, Claude, the eleven independently aligned ELEPHANT models, and the separate model set in the Pangram benchmark makes it unlikely to be an idiosyncrasy of any one system.

### 4.5. Conclusion

Large language models use the language of obligation differently from contemporary humans, and they do so in a patterned way that reflects a combination of formal-register exposure, procedural training data, and instruction-tuned preferences for measured advisory language. The deontic gap between AI and human writers is not random noise; it concentrates on the constructions that have become most characteristic of informal contemporary human writing (*should*, *have to*, *had to*, contractions) while sparing the constructions that are overrepresented in instructional and procedural text (*need to*, *cannot*). Situating AI modal rates against the Google Books historical record as a heuristic calibration reveals that AI fiction completions fall in the range associated with mid-to-late twentieth-century published prose, while contemporary human fiction writers use deontic constructions at rates that exceed the historical book baseline. The pattern is best understood not as temporal displacement but as register fixation: LLMs have learned from formal text, and the deontic language of formal text is not the deontic language of contemporary informal human life. This paper establishes that descriptive pattern and offers a plausible interpretive framework. We began by asking why an AI–human deontic gap might influence normative uptake in humans; we have now established empirically that such a gap exists, which opens a new agenda for research into the extent to which it matters in practice, for example, for how readers interpret AI-generated guidance. Recent evidence that LLM-favored lexical patterns are entering human speech (Yakura et al., 2024) suggests that AI-generated language may itself become part of the linguistic environment from which future speakers draw. Whether the deontic profile identified here remains a marker of machine text or begins to influence how humans themselves formulate obligation is an open question for longitudinal pragmatics.

---

### Declaration of Competing Interests

The authors declare no competing interests.

### Declaration of Generative AI in the Research and Writing Process

This study concerns the language produced by large language models, and AI-generated text is therefore part of the object of study. The AI-authored corpora analyzed here (GPT-4o, Claude, and the eleven models in the ELEPHANT benchmark) are described in the Method section.

Separately, the authors used AI-assisted tools in conducting and reporting the research. Anthropic's Claude (via the Claude Code interface) was used to help write and debug the R analysis pipeline, to generate the figures from author-specified designs, and to assist in drafting and copy-editing the manuscript. All analytic decisions, corpus-construction choices, statistical specifications, and interpretations were made and verified by the authors, who take full responsibility for the accuracy of the reported numbers, the appropriateness of the analyses, and the content of the manuscript. Every headline statistic in the manuscript was regenerated and checked against the analysis outputs by the authors, and no citations, data, or findings were accepted from an AI tool without independent verification.

### Author Contributions (CRediT)

**Daniel Hart:** Conceptualization, Methodology, Formal analysis, Writing – original draft, Writing – review & editing, Supervision. **Sarah Allred:** Conceptualization, Methodology, Formal analysis, Writing – review & editing. **Joseph Abbas:** Conceptualization, Formal analysis, Writing – review & editing. **Morenike Alugo:** Conceptualization, Formal analysis, Writing – review & editing.

### Funding

Morenike Alugo was supported by a grant from the Rutgers Democracy Lab.

### Data and Code Availability

The Pangram benchmark used in the out-of-sample replication is publicly available at https://checkforai-public.s3.amazonaws.com/benchmark.csv (Emi & Spero, 2024). The WritingPrompts source dataset is available at https://huggingface.co/datasets/euclaise/writingprompts (Fan et al., 2018). The gsingh dataset is available at https://huggingface.co/datasets/gsingh1-py/train. The HC3 corpus is available at https://huggingface.co/datasets/Hello-SimpleAI/HC3 (Guo et al., 2023). The Google Books Ngram data were retrieved via the ngramr R package. All analysis code and derived data needed to reproduce the reported tables and figures are openly available at https://doi.org/10.17605/OSF.IO/YCJWH.

---

---

## Supplementary Material

*Table S1*

*Positive Deontic-Modal Rates and Length-Adjusted Rate Ratios for Human Writers and Eleven Language Models in the ELEPHANT Advice Corpus*

| Responder | OEQ rate (per 10k) | OEQ rate ratio [95% CI] | AITA rate (per 10k) | AITA rate ratio [95% CI] |
|---|---|---|---|---|
| Human | 47.9 | (reference) | 51.0 | (reference) |
| GPT-4o | 14.6 | 0.30 [0.28, 0.32] | 13.9 | 0.27 [0.24, 0.30] |
| GPT-5 | 17.1 | 0.35 [0.33, 0.38] | 18.9 | 0.37 [0.33, 0.41] |
| Claude | 18.8 | 0.38 [0.35, 0.42] | 18.0 | 0.35 [0.31, 0.40] |
| Gemini | 23.1 | 0.47 [0.44, 0.51] | 33.5 | 0.66 [0.59, 0.73] |
| DeepSeek | 22.3 | 0.46 [0.42, 0.49] | 21.7 | 0.42 [0.38, 0.47] |
| Qwen | 17.8 | 0.36 [0.33, 0.39] | 19.5 | 0.38 [0.34, 0.43] |
| Llama-8B | 13.0 | 0.26 [0.24, 0.28] | 22.7 | 0.44 [0.40, 0.50] |
| Llama-17B | 13.3 | 0.27 [0.25, 0.29] | 15.9 | 0.31 [0.27, 0.35] |
| Llama-70B | 15.0 | 0.30 [0.28, 0.33] | 18.7 | 0.37 [0.33, 0.41] |
| Mistral-7B | 13.8 | 0.28 [0.26, 0.31] | 11.9 | 0.23 [0.20, 0.26] |
| Mistral-24B | 22.3 | 0.45 [0.42, 0.48] | 23.7 | 0.46 [0.41, 0.52] |

*Note.* Rates are counts per 10,000 words, pooled over responses of at least 50 words. OEQ = open-ended personal-advice questions; AITA = r/AmItheAsshole posts (crowd verdict "you're the asshole"). The rate ratio is the multiplicative difference from the human rate estimated by a Poisson generalized linear mixed model fit separately for each model, with a log word-count offset and a prompt-level random intercept, so that it adjusts for the greater length of machine-generated responses. A ratio below 1.00 indicates that the model marks positive obligation less often than human writers. AI responses are from the ELEPHANT dataset (Cheng et al., 2026) and were generated by the original authors rather than under matched decoding conditions.

---

*Table S2*

*Single-Feature Classification Performance (Positive-Modal Rate) on the Pangram Benchmark, by Domain*

| Domain | *n* AI | *n* Human | AUC |
|---|---|---|---|
| Creative writing | 128 | 150 | .61 |
| Student writing | 128 | 150 | .62 |
| Scientific | 128 | 149 | .59 |
| News | 128 | 148 | .58 |
| Reviews | 128 | 132 | .52 |
| Blog | 80 | 49 | .67 |
| Encyclopedic | 64 | 94 | .49 |
| Books | 48 | 50 | .57 |
| Q&A | 48 | 41 | .57 |
| Email | 48 | 34 | .60 |
| **Overall** | **928** | **997** | **.57** |

*Note.* Values are the area under the ROC curve (AUC) for positive-modal rate as a single-feature classifier of the human/AI label, with the human label as the positive class (.50 = no separation; 1.00 = perfect separation). The overall AUC of .57 indicates that positive-modal rate distinguishes individual human and AI documents only weakly, even though the population-level direction is consistent across domains. Per-domain positive-modal rates are shown in Figure 2. Eight LLMs contributed to the AI documents (including GPT-3.5, GPT-4, Gemini Pro, Mistral 7B, and LLaMA 2).

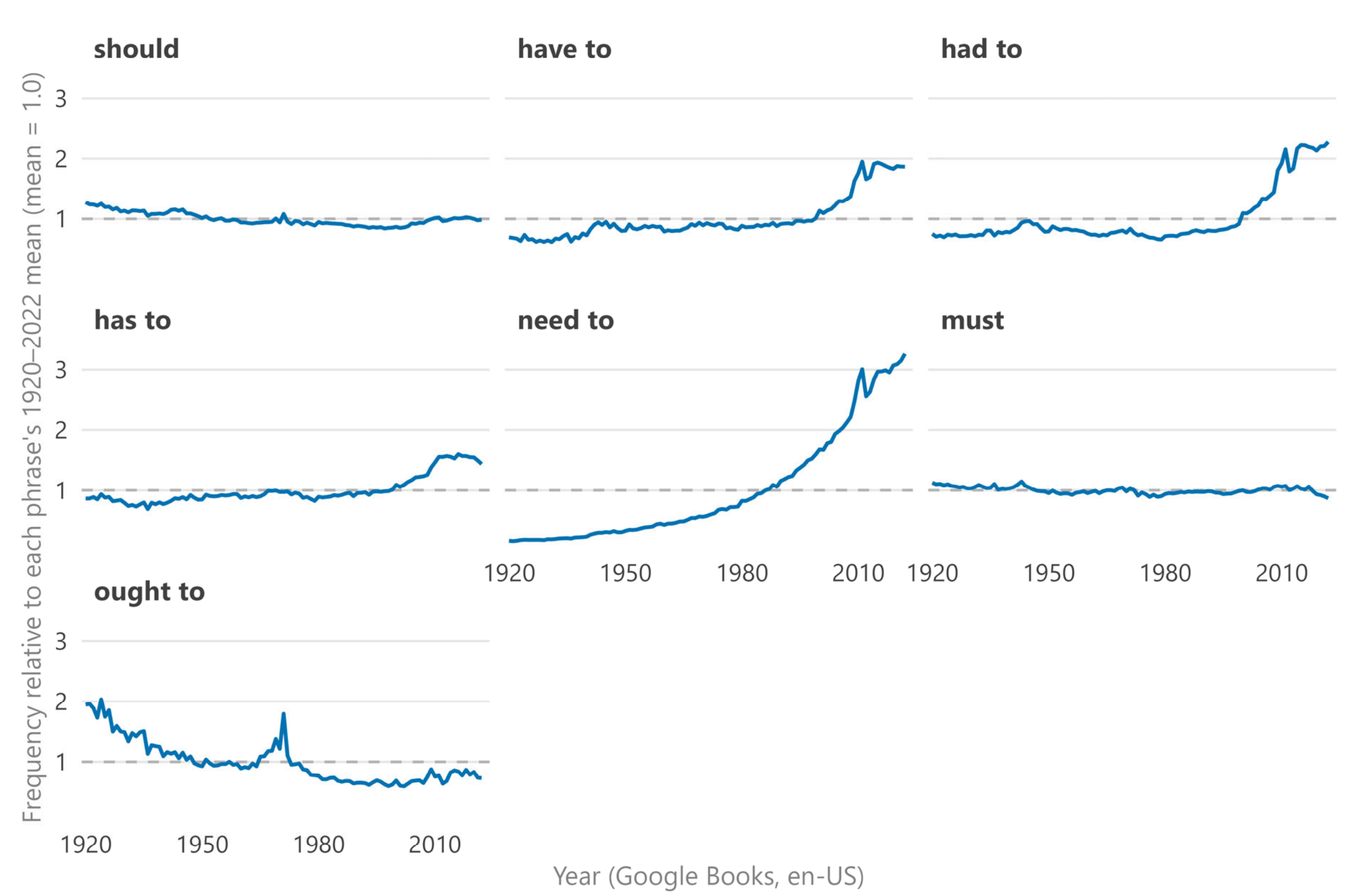

*Figure S1.* Historical trajectory of each positive deontic construction in Google Books (en-US, 1920–2022), with each series indexed to its own 1920–2022 mean (dashed line at 1.0) so that constructions with very different base rates can be compared on a single scale. *Need to* rises steadily across the century, whereas *have to*, *had to*, and *has to* are largely flat before rising in recent decades, and *should* and *must* are essentially flat. These non-monotonic paths are why the historical-change analysis (Figure 5) treats the 1960-to-2019 difference as a descriptive summary rather than a precise estimate.

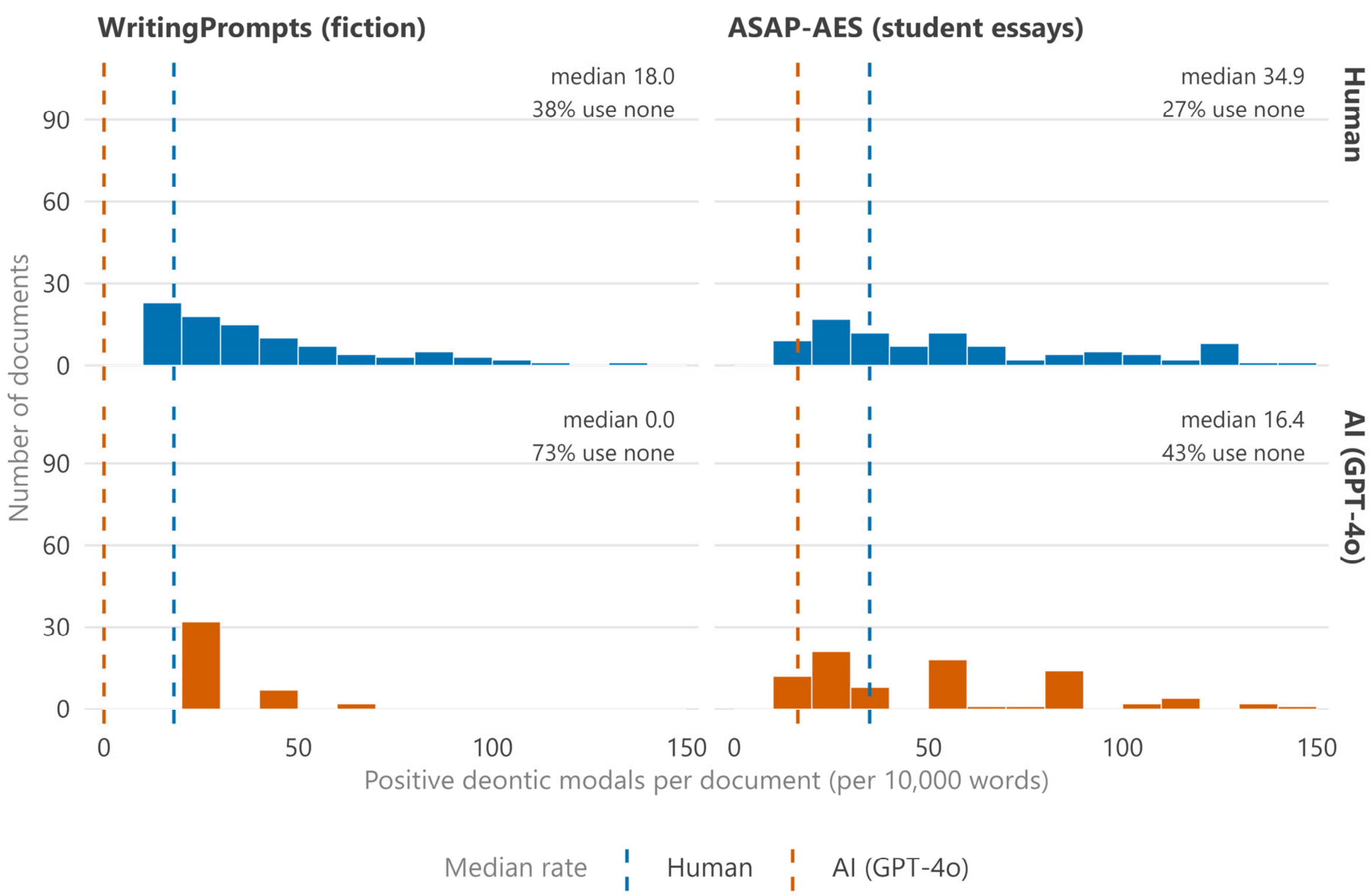

*Figure S2.* Distribution of the per-document positive deontic-modal rate (per 10,000 words) for human and GPT-4o texts in the two prompt-matched corpora, WritingPrompts (fiction) and ASAP-AES (student essays), on documents of at least 50 words. Dashed lines mark the median rate for each writer (blue = human, vermillion = GPT-4o), drawn in both rows of each corpus so that the two medians can be compared directly; annotations give each panel's median and the percentage of documents that contain no positive deontic modal. In both corpora the human–AI gap is carried by the right tail of the human distribution rather than by a uniform shift, and AI documents pile up at zero more often than human documents. The horizontal axis is truncated at 150 per 10,000 words; a small number of higher-rate documents are omitted from the display but retained in all statistics.